\def\year{2022}\relax
\documentclass[letterpaper]{article} 
\usepackage{aaai22}  
\usepackage{times}  
\usepackage{helvet}  
\usepackage{courier}  
\usepackage[hyphens]{url}  
\usepackage{graphicx} 
\usepackage{natbib}  
\usepackage{caption} 
\usepackage{color}
\usepackage{siunitx}
\usepackage{bm}
\usepackage{multirow}
\usepackage{dsfont}
\usepackage{tabularx}
\usepackage{caption}
\usepackage{verbatim}
\usepackage{float}
\usepackage{pifont}
\usepackage{amsthm}
\usepackage{verbatim}
\usepackage{rotating} 
\usepackage{tabularx}
\usepackage{amsmath}
\usepackage{amssymb}
\usepackage{booktabs}
\usepackage{mathrsfs}
\DeclareCaptionStyle{ruled}{labelfont=normalfont,labelsep=colon,strut=off} 
\usepackage{algorithm}
\usepackage{algorithmic}

\usepackage{newfloat}
\usepackage{listings}
\floatstyle{ruled}
\newfloat{listing}{tb}{lst}{}
\floatname{listing}{Listing}
\title{Reliable Inlier Evaluation for Unsupervised Point Cloud Registration}
\author{
	Yaqi Shen,
	Le Hui,
	Haobo Jiang,
	Jin Xie\thanks{Corresponding author.},
	Jian Yang
}
\affiliations{
   	PCA Lab, Key Lab of Intelligent Perception and Systems for High-Dimensional Information of Ministry of Education\\
   	Jiangsu Key Lab of Image and Video Understanding for Social Security\\
   	School of Computer Science and Engineering, Nanjing University of Science and Technology, Nanjing, China\\
   \{syq, le.hui, jiang.hao.bo, csjxie, csjyang\}@njust.edu.cn
}

\usepackage{bibentry}

\begin{document}
	\maketitle
	
	\begin{abstract}	
	Unsupervised point cloud registration algorithm usually suffers from the unsatisfied registration precision in the partially overlapping problem due to the lack of effective inlier evaluation. In this paper, we propose a neighborhood consensus based reliable inlier evaluation method for robust unsupervised point cloud registration. It is expected to capture the discriminative geometric difference between the source neighborhood and the corresponding pseudo target neighborhood for effective inlier distinction. Specifically, our model consists of a matching map refinement module and an inlier evaluation module. In our matching map refinement module, we improve the point-wise matching map estimation by integrating the matching scores of neighbors into it. The aggregated neighborhood information potentially facilitates the discriminative map construction so that high-quality correspondences can be provided for generating the pseudo target point cloud. Based on the observation that the outlier has the significant structure-wise difference between its source neighborhood and corresponding pseudo target neighborhood while this difference for inlier is small, the inlier evaluation module exploits this difference to score the inlier confidence for each estimated correspondence. In particular, we construct an effective graph representation for capturing this geometric difference between the neighborhoods. Finally, with the learned correspondences and the corresponding inlier confidence, we use the weighted SVD algorithm for transformation estimation. Under the unsupervised setting, we exploit the Huber function based global alignment loss, the local neighborhood consensus loss, and spatial consistency loss for model optimization. The experimental results on extensive datasets demonstrate that our unsupervised point cloud registration method can yield comparable performance. Our code is available at \emph{\url{https://github.com/supersyq/RIENet}}.
	\end{abstract}
	
	\section{Introduction}
	Point cloud registration, as a fundamental and important topic in the 3D vision field, has been widely applied in various vision tasks, including 3D scene reconstruction \cite{agarwal2011building,schonberger2016structure}, object pose estimation \cite{wong2017segicp}, and SLAM \cite{deschaud2018imls,zhang2014loam}. 
	The goal of the point cloud registration is to align point clouds by estimating the rigid transformation between them.
	Due to large pose changes and occlusions, point cloud registration is still a challenging problem in real applications. 
	
	In recent years, owing to the discriminative representation ability of deep learning, deep point cloud registration methods have achieved increasing research interests.
	Most of these methods \cite{pais20203dregnet,fu2021robust, ali2021rpsrnet} focus on learning the rigid transformation in a supervised manner, where a large number of ground truth transformations are required as the supervision signal for model training. 
	However, collecting required ground truth transformations is expensive and time-consuming, which may greatly increase the training cost and hinder their applications in the real world. 
	To handle the concerns above, much more effort has been dedicated to the unsupervised deep point cloud registration field. 
	For example, alignment error is a commonly used optimization signal for unsupervised training \cite{wang2020unsupervised}. 
	By minimizing the alignment error (e.g., Chamfer distance) between the transformed source point clouds and the target point clouds, it is expected to learn the optimal rigid transformation so that the point cloud pair can be aligned perfectly. 
	Nevertheless, for the partially overlapping point cloud pair, due to the existing noise error from the outliers, its optimization is prone to stick into the local optima.
	In addition, the cycle consistency based methods \cite{yang2020mapping} perform unsupervised optimization by constructing a cycle constraint as the self-supervised signal. 
	However, the cycle consistency based loss still suffers from the outlier dilemma since the outliers may not be able to form a closed loop well. 
	
	In this paper, we propose a novel neighborhood consensus based reliable inlier evaluation framework for unsupervised deep point cloud registration. 
	By constructing an effective graph representation for the source neighborhood and the learned pseudo target neighborhood of a point, it aims to exploit the structure-wise difference between the source and the pseudo target neighborhoods to identify whether this point is an inlier or not. 
	Specifically, our model mainly consists of two modules, including a matching map refinement module and an inlier evaluation module. 
	Given a point cloud pair with unknown correspondence, the matching map refinement module aims to provide a high-quality correspondence estimation for constructing a pseudo target point cloud. 
	Since the locally similar single points may confuse the point-wise matching map for incorrect correspondence estimation, we improve the matching map by aggregating the neighborhood scores into the score of the center point.
	The rich neighborhood information can effectively facilitate the discriminative matching map construction and thus the high-quality correspondences can be provided.
	With the constructed pseudo target point cloud using the learned correspondences, the inlier evaluation module further exploits the geometric difference between the source neighborhood and the corresponding pseudo target neighborhood for reliable inlier evaluation. 
	The main motivation is based on the observation that the inlier in the source point cloud tends to have a similar neighborhood structure as its corresponding pseudo neighborhood in the pseudo target point cloud while the outlier may have a significant structural difference. Therefore, we utilize the structure difference as the metric for the inlier confidence estimation. 
	Particularly, we construct a learnable and effective graph representation to capture this structure difference between the two neighborhoods.
	Finally, with the learned correspondences and the corresponding inlier confidence scores, we utilize the weighted Singular Value Decomposition (weighted SVD) for the rigid transformation estimation.
	In the unsupervised setting, instead of using the ground truth transformation, we design a Huber function based robust alignment loss, a local neighborhood consensus loss, and a spatial consistency loss for model training. 
	We evaluate our method on extensive benchmark datasets, including ModelNet40 \cite{wu20153d}, 7Scenes \cite{shotton2013scene}, ICL-NUIM \cite{choi2015robust}, and KITTI \cite{geiger2012we} and the experimental results verify the effectiveness of our method.
	
	To summarize, our main contributions are as follows: 
	\begin{itemize}
		\item  We propose a novel reliable inlier evaluation framework for unsupervised point cloud registration, where the graph-structure difference between the neighborhoods is used to identify the inlier.
		\item We enhance the original point-wise matching map with the neighborhood consensus for high-quality correspondence generation.
		\item Extensive experimental results on the different benchmark datasets demonstrate that our method can achieve superior registration precision.
	\end{itemize}
	
	\section{Related Work}
	\noindent \textbf{Traditional point cloud registration.} ICP \cite{besl1992method} iteratively modifies the transformation to minimize the error between the corresponding points. The main drawback of ICP is that it requires a good initialization to prevent the model from getting a local optimal. Therefore, several ICP variants \cite{phillips2007outlier,bouaziz2013sparse,segal2009generalized} have been proposed, which are also prone to getting stuck into the local minima. To obtain the global optimal solution, Go-ICP \cite{yang2013go} is developed with a branch-and-bound optimization, but it is also sensitive to outliers due to the least-squares objective. 
	In addition, the randomized methods \cite{chen1999ransac,le2019sdrsac} using different sampling strategies are proposed for point cloud registration. 
	Among them, Super4PCS \cite{mellado2014super}, a global registration method, reduces the quadratic complexity of congruent set extraction in 4PCS \cite{aiger20084} to the linear. Besides, some hand-crafted descriptors \cite{rusu2008aligning, rusu2009fast, tombari2010unique} based methods are proposed to estimate rigid transformation with RANSAC \cite{fischler1981random}.
	
	\begin{figure*}[t]
		\begin{center}
			\includegraphics[width=\linewidth]{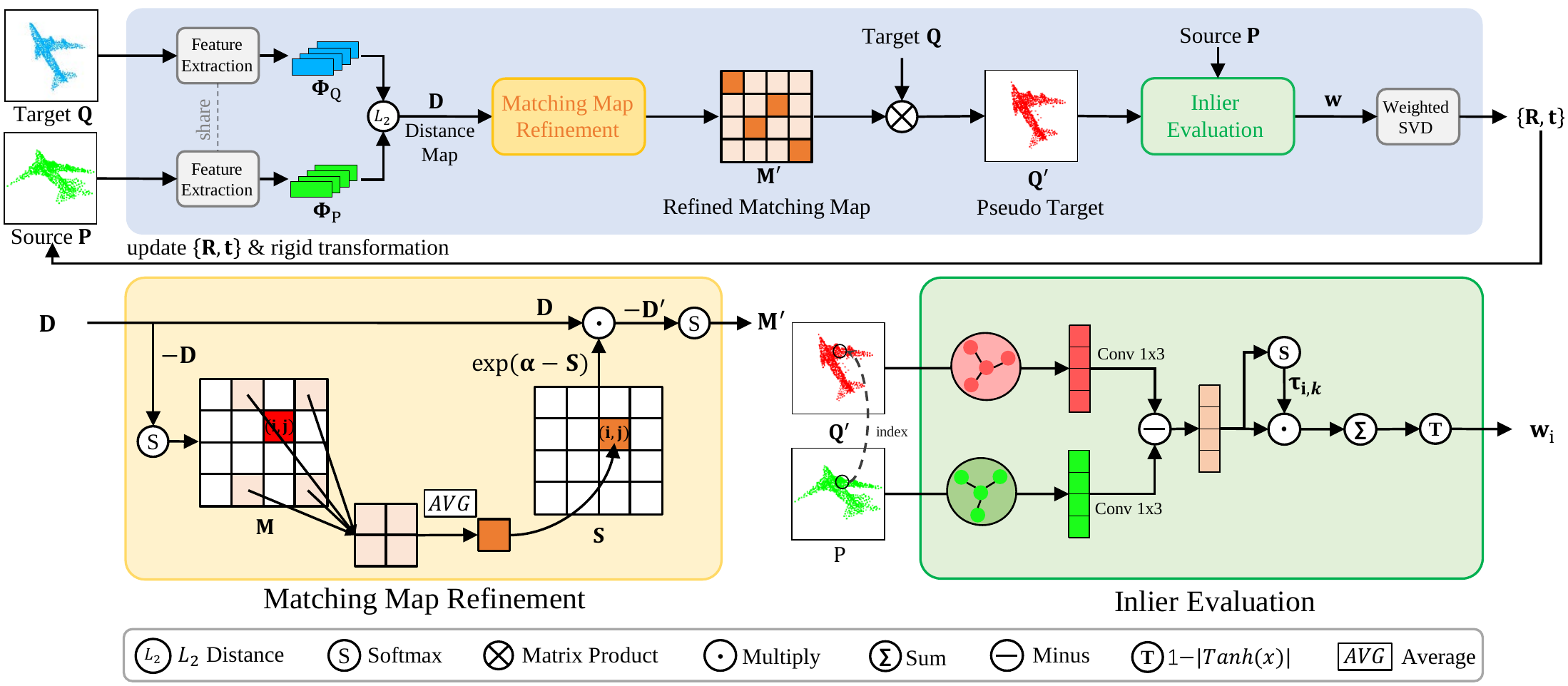}
		\end{center}
		\vskip -10pt
		\caption{An overview of our proposed method. We first extract the local features of point clouds. Then we generate the point-wise matching map $\mathbf{M}$ and refine it with neighborhood consensus to predict the pseudo matching points for constructing the pseudo target point cloud $\mathbf{Q}'$. Based on the pseudo target point cloud $\mathbf{Q}'$ and the source point cloud $\mathbf{P}$, we adopt an inlier evaluation module based on neighborhood consensus, which can output the confidence $\mathbf{w}_i$ of each pseudo matching pair $\{\mathbf{p}_i,\mathbf{q}^{\prime}_i\}$. Finally, we feed the correspondences along with the weight $\mathbf{w}$ into the Weighted SVD to obtain the rigid transformation.}
		\label{model}
	\end{figure*}
	
	\noindent \textbf{Learning-based point cloud registration.} Recently, deep learning has been successfully applied to the point cloud registration which can get the rigid transformation in an end-to-end manner. There are some supervised methods \cite{lu2019deepvcp,yuan2020deepgmr,sarode2020masknet,huang2021predator}. PointNetLK~\cite{aoki2019pointnetlk} modifies the Lucas $\&$ Kanada (LK) algorithm and integrates it into the PointNet \cite{qi2017pointnet} for point cloud registration. Following PointNetLK, a voxelized analytical PointNetLK \cite{li2021pointnetlk} is proposed to circumvent the numerical instabilities of the original PointNetLK. 
	Following the soft matching map based methods \cite{wang2019deep,wang2019prnet}, RPM-Net \cite{yew2020rpm} uses the Sinkhorn layer to enforce the doubly stochastic constraints on the matching map for reliable correspondences. \cite{li2020iterative} proposes an iterative distance-aware similarity matrix convolution network with two-stage point elimination technique for point cloud registration. 
	Besides, there are some graph neural network based methods \cite{fischer2021stickypillars, fu2021robust}, which achieve robust feature matching via the self/cross attention for context aggregation on point clouds. 
	Moreover, other learning-based supervised methods \cite{sarode2019pcrnet,gojcic2020learning,choy2020deep,bai2021pointdsc} also achieve impressive performance.
	
	Additionally, there are some end-to-end unsupervised point cloud registration methods \cite{kadam2020unsupervised,feng2021recurrent,li2019pc,el2021unsupervisedr}. 
	\cite{yang2020mapping,groueix2019unsupervised} employ cycle consistency across the pairwise point clouds for points matching, which cannot be trained directly on the partial data. And the global feature based FMR \cite{huang2020feature} uses point cloud reconstruction for feature extraction with poor registration precision on partial/noisy data. \cite{li2020unsupervised} proposes the model of shape completion and registration, but this model needs to be trained separately on the point cloud of each category with a latent code. 
	Besides, some deep unsupervised learning based local descriptors of point clouds are developed for point cloud correspondence \cite{li2021updesc,li2019usip}. Those methods cannot obtain the rigid transformation in an end-to-end manner, they focus on feature extractor followed by the RANSAC for geometric matching. 
	Besides, \cite{bauer2021reagent,jiang2021CEMNet,jiangplanning} treat the point cloud registration as a reinforcement learning task.

	\section{Method}
	In this section, we demonstrate our neighborhood consensus based unsupervised point cloud registration framework in detail.
	As shown in Figure~\ref{model}, it iteratively performs the matching map refinement and the inlier evaluation modules. Thus, the high-quality correspondence estimation and the reliable inlier confidence evaluation can be learned for robust rigid transformation estimation. 
	
	\subsection{Matching Map Refinement Module}\label{matching} 
	\label{refine}
	The matching map based point cloud registration methods \cite{yew2020rpm} usually exploits the single-point feature distance as the matching score for map construction. However, the non-corresponding points with similar point features potentially may mislead those methods for wrong correspondence estimation. To relieve it, the matching map refinement module targets at clarifying the ambiguous single-point feature by merging the discriminative matching scores of the neighboring points into it for reliable correspondence identification. 
	
	Given the source point cloud $\mathbf{P}=\{\mathbf{p}_i\in\mathbb{R}^3\mid i=1,...,N\}$ and the target point cloud $\mathbf{Q}=\left\{\mathbf{q}_j\in\mathbb{R}^3 \mid j=1,...,M\right\}$, we first extract their local descriptors ${\Phi}_{\mathbf{P}}\in\mathbb{R}^{N\times d}$ and ${\Phi}_{\mathbf{Q}}\in\mathbb{R}^{M\times d}$ with the Dynamic Graph CNN (DGCNN) \cite{wang2019dynamic}. Then, for each point cloud pair $\mathbf{p}_i$ and $\mathbf{q}_j$, we calculate their point-wise matching score with the normalized negative feature distance as below:
	\begin{equation}\label{M1}
	\mathbf{M}_{i,j}= \operatorname{softmax} \left(\left[-\mathbf{D}_{i,1},\ldots, -\mathbf{D}_{i,M}\right]\right)_{j},
	\end{equation}
	where $\mathbf{D}_{i,j} = \left\|{\Phi}_{\mathbf{p}_i} - {\Phi}_{\mathbf{q}_j}\right\|_2$ denotes the Euclidean distance between the single-point features of points $\mathbf{p}_i$ and $\mathbf{q}_j$. 
	Based on the estimated point-wise matching score, their neighborhood-wise score can be calculated by averaging the correspondence scores of its surrounding points: 
	\begin{equation}\label{eq1}
	\mathbf{S}_{i,j}= \frac{1}{K}\sum_{\mathbf{p}_{i'} \in \mathcal{N}_{\mathbf{p}_i}}\sum_{\mathbf{q}_{j'} \in \mathcal{N}_{\mathbf{q}_j}}\mathbf{M}_{i',j'},
	\end{equation}
	where $\mathcal{N}_{\mathbf{p}_i}$ denotes the $k$-nearest neighbor points around the point $\mathbf{p}_i$. 
	The high neighborhood score means the surrounding points consistently tend to have large matching probabilities and vice versa. 
	Based on the fact that the real correspondence usually owns the high neighborhood similarity while this similarity for incorrect correspondence is prone to be low,
	we can effectively exploit this neighborhood score to identify the non-corresponding point pair.
	Finally, the refined matching map can be formulated as:
	\begin{equation}\label{eq0}
	\begin{split}
	\mathbf{M}'_{i,j}&= \operatorname{softmax} \left(\left[-\mathbf{D}'_{i,1},\ldots, -\mathbf{D}'_{i,M}\right]\right)_{j},\\
	\mathbf{D}{'}_{i,j} &= \exp \left(\mathbf{\alpha} - \mathbf{S}_{i,j}\right) * \mathbf{D}_{i,j},
	\end{split}
	\end{equation}
	where the refined feature distance $\mathbf{D}'_{i,j}$ is negatively related to the neighborhood score  $\mathbf{S}_{i,j}$ and we design an exponential function to control the changing ratio. The hyper-parameter $\alpha$ controls the influence of the neighborhood consensus. Based on the refined matching map $\mathbf{M}' \in \mathbb{R}^{N \times M}$, for each point $\mathbf{p}_i$ in the source point cloud, we utilize a point predictor to generate a pseudo correspondence $\mathbf{q}'_i$ as following:
	\begin{equation}\label{predictor}
	\begin{split}
	\mathbf{q}'_i = \sum_{j=1}^M \mathbf{M}'_{i,j} \cdot \mathbf{q}_j \in \mathbb{R}^3.
	\end{split}
	\end{equation} 
	Notably, if $\mathbf{p}_i$ is an inlier, owing to the high-quality matching map estimation, the predicted pseudo correspondence usually can obtain a correct location (i.e., $\mathbf{q}'_i$ approaches to the real correspondence of $\mathbf{p}_i$). However, if $\mathbf{p}_i$ is an outlier, it may have a low matching probability with each target point and the resulting pseudo point tends to have an incorrect and unstable location. We use this observation for inlier confidence evaluation and more details can be seen in the following section. 
	
	\subsection{Inlier Evaluation Module}
	\label{inlier}
	\begin{figure}[t]
		\begin{center}
			\includegraphics[width=0.8\linewidth]{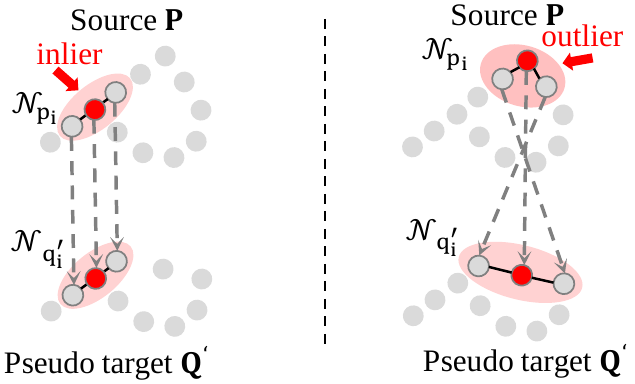}
		\end{center}
		\vskip -10pt
		\caption{The examples of inlier and outlier.}
		\label{nn}
		\vskip -10pt
	\end{figure}
	
	In this section, we aim to exploit the geometric difference between the source neighborhood and the corresponding pseudo target neighborhood for effective inlier evaluation on the learned correspondences $\left\{\left(\mathbf{p}_i,\mathbf{q}'_i\right)\right\}$. 
	The main motivation for our inlier distinction is presented below. If the point $\mathbf{p}_i$ in the source point cloud is a reliable inlier, its surrounding points (source neighborhood $\mathcal{N}_{\mathbf{p}_i}$) also tend to be the inliers. Thus, benefitting from the learned high-quality matching map, their resulting pseudo correspondences (pseudo target neighborhood $\mathcal{N}_{\mathbf{q}'_i}$) can have a similar geometric structure as the source neighborhood. 
	Instead, if $\mathbf{p}_i$ is an outlier, its neighbors may also be the outliers. Since the unstable location of pseudo target points for outliers as described before, the generated pseudo target neighborhood may be chaotic and its spatial structure is prone to be significantly different from the source neighborhood.
	Based on the discussion above, we formulate this structure difference as the inlier confidence estimation for reliable inlier selection. The visualization examples can be seen in Figure \ref{nn}.
	
	By constructing a learnable graph representation on the neighborhood, the inlier evaluation module targets at adaptively capturing the structure-wise difference between the neighborhoods $\mathcal{N}_{\mathbf{p}_i}$ and $\mathcal{N}_{\mathbf{q}^{\prime}_i}$ for inlier evaluation.
	The edge features of the neighborhoods $\mathcal{N}_{\mathbf{p}_i}$ and $\mathcal{N}_{\mathbf{q}^{\prime}_i}$ are defined as $\mathbf{e}^{p}_{i,k} = \mathbf{p}_i - \mathbf{p}_k$ and $\mathbf{e}^{q^{\prime}}_{i,k} = \mathbf{q}^{\prime}_i - \mathbf{q}^{\prime}_k$, respectively, where $\mathbf{p}_k$ and $\mathbf{q}'_k$ are the points inside the neighborhoods $\mathcal{N}_{\mathbf{p}_i}$ and $\mathcal{N}_{\mathbf{q}'_i}$. 
	In order to better capture the relevance between the adjacent points and promote their message propagation, for each point, we fuse the edge features of its adjacent points into it through an EdgeConv $v$ with large ($1\times 3$) convolutional kernel size \cite{wang2019dynamic}. 
	It is noted that due to the consistent order of the points in the neighborhood, we can directly employ the EdgeConv $v$ on the adjacent points for feature fusion.
	In detail, for each point $\mathbf{p}_i\in \mathbf{P}$ and $\mathbf{q}'_i\in\mathbf{Q}'$, we use the subtraction of the fused edge features as following to characterize the difference between the neighborhoods:
	\begin{equation}\label{e0}
	\mathbf{e}_{i,k} = v \left(\mathbf{e}^{p}_{i,k}\right) - v \left(\mathbf{e}^{q^{\prime}}_{i,k}\right).
	\end{equation}
	Next, with the edge difference between the two neighborhoods, we further adaptively learn the attention coefficients $\bm{\tau}_{i,k}$ of each edge via a softmax function:
	\begin{equation}\label{eq8}
	\bm{\tau}_{i,k} = \operatorname{softmax} \left(\left[u\left(\mathbf{e}_{i,1}\right),...,u\left(\mathbf{e}_{i,K}\right)\right]\right)_k,
	\end{equation}
	where $K$ is the number of the neighbors and $u$ is another EdgeConv.
	Finally, we sum the edge difference weighted by the attention coefficients to learn the inlier confidence $\mathbf{w}_{i}$ of the pseudo correspondence $\{\mathbf{p}_i,\mathbf{q}'_i\}$ as below:
	\begin{equation}\label{eq9}
	\mathbf{w}_{i} = 1 - \operatorname{tanh}\left(\left|g \left(\sum\nolimits_{k=1}^{K}\mathbf{\tau}_{i,k} * \mathbf{e}_{i,k}\right) \right|\right),
	\end{equation} 
	where $g$ is a unary function.
	
	\subsection{Loss Function}\label{loss1}
	In each iteration, we can obtain a set of pseudo correspondences and their corresponding inlier confidence $\mathbf{w}$. Then, we employ the weighted SVD on the pseudo correspondences with weights $\mathbf{w}$ to estimate the rigid transformation $\left\{\mathbf{R},\mathbf{t}\right\}$. Under the unsupervised setting, instead of using the ground truth transformations, we construct the following unsupervised loss functions for model optimization.
	
	\noindent \textbf{Global alignment loss.} 
	We first exploit the alignment error based loss function to train our model. To handle the partially-overlapping problem well, we integrate the Huber function, a robust loss function insensitive to the outliers, as below into our alignment loss: 
	\begin{equation}\label{huber}
	\ell_{\beta}(u)= \begin{cases}\frac{1}{2} u^{2}, & \text { if }|u| \leq \beta \\ \beta\left(|u|-\frac{1}{2} \beta\right), & \text { otherwise }\end{cases},
	\end{equation}
	where the hyper-parameter $\beta$ controls the range of the inlier and the alignment error from the outlier beyond this range just can obtain the sublinear optimization. 
	The final Huber function based global alignment loss can be formulated as:
	\begin{equation}
	\begin{split}
	\mathcal{L}_g\left(\mathbf{P}^{\prime}, \mathbf{Q}\right) =&\sum_{\mathbf{p}' \in \mathbf{P}^{\prime}} \ell_{\beta}\left(\min _{\mathbf{q} \in \mathbf{Q}}\|\mathbf{p}^{\prime}-\mathbf{q}\|_{2}^{2}\right) \\
	+&\sum_{\mathbf{q} \in \mathbf{Q}} \ell_{\beta}\left(\min _{\mathbf{p}' \in \mathbf{P}^{\prime}}\|\mathbf{q}-\mathbf{p}'\|_{2}^{2}\right),
	\end{split}
	\end{equation}
	where $\mathbf{P}'$ denotes the transformed source point cloud with the estimated transformation. 
	However, although our robust function based alignment loss can relieve the interference of the outliers to some extent, these outliers may still potentially mislead the model optimization to local minima dilemma. To this end, we propose the reliable inliers based neighborhood consensus loss and the spatial consistency loss to further eliminate the influence from the outliers.
	
	\noindent \textbf{Neighborhood consensus loss.} 
	With the learned confidence $\mathbf{w}$, we can obtain the reliable inliers of the source and target point clouds by taking the $k^\prime$ point pairs with the highest weights. 
	We denote the inliers of the source and target point clouds as $\mathbf{X}\in \mathbb{R}^{k^{\prime}\times3}$ and $\mathbf{Y}\in \mathbb{R}^{k^{\prime}\times3}$, respectively. 
	Then, the neighborhood consensus loss uses the neighborhood consistency between the inliers as the self-supervised signal for model optimization. 
	Ideally, for each inlier ${\mathbf{x}_i \in \mathbf{X}}$, after performing transformation $\left\{\mathbf{R},\mathbf{t}\right\}$ on its neighborhood $\mathcal{N}_{\mathbf{x}_i}$ ($k$ nearest neighbor points), the transformed $\mathcal{N}_{\mathbf{x}_i}$ can align the neighborhood $\mathcal{N}_{\mathbf{y}_i}$ of its correspondence $\mathbf{y}_i$ perfectly. 
	Thus, the neighborhood consensus loss is defined as:
	\begin{equation}
	\begin{split}
	 \mathcal{L}_{n} = \sum_{\mathbf{x}_i \in \mathbf{X}, \mathbf{y}_i \in \mathbf{Y}} \sum_{\mathbf{p}_j \in \mathcal{N}_{\mathbf{x}_i}, \mathbf{q}_j \in \mathcal{N}_{\mathbf{y}_i} } \|\mathbf{R}\mathbf{p}_j + \mathbf{t} - \mathbf{q}_j \|_{2}.	
	\end{split}
	\end{equation}
	
	\noindent \textbf{Spatial consistency loss.} 
	In order to further eliminate the spatial gap between the pseudo target correspondence and the real target correspondence for each selected inlier $\mathbf{x}_i \in \mathbf{X}$, we exploit the cross-entropy based spatial consistency loss to sharpen their matching distributions as below:
	\begin{equation}\label{eq}
	\mathcal{L}_s = -\frac{1}{|\mathbf{X}|}\sum_{\mathbf{x}_i\in\mathbf{X}}\sum_{j=1}^M \mathbb{I}\left\{j=\arg\max_{j'}\mathbf{M}'_{i,j'}\right\}\log\mathbf{M}'_{i,j},
	\end{equation}
	where $\mathbb{I}\left\{\cdot\right\}$ denotes the indicator function and we use the target point $\mathbf{q}_{j^*}$ with the largest matching probability to estimate the real target correspondence. By improving the matching probability of the ``real'' target correspondence ($\mathbf{M}'_{i,j^*}\rightarrow 1$), the resulting pseudo target correspondence via Eq. \ref{predictor} tends to further spatially approach to the ``real'' target correspondence $\mathbf{q}_{j^*}$ well.  
	Finally, we utilize a comprehensive loss function as below to optimize our model:
	\begin{equation}
	\mathcal{L} = \mathcal{L}_g + \gamma * \mathcal{L}_{n} + \theta * \mathcal{L}_{s},
	\end{equation}
	where $\gamma$ and $\theta$ are the hyper-parameters to control the weights of the used neighborhood consensus loss and the spatial consistency loss.
	
	\begin{table*}[t]
		\setlength{\abovecaptionskip}{0pt}
		\setlength{\belowcaptionskip}{0pt}
		\begin{center}
			{    \resizebox{1\linewidth}{!}{
					\begin{tabular}{lcccc|cccc|cccccccc}
						\toprule
						\multicolumn{1}{c}{\multirow{2}{*}{Model}} &\multicolumn{4}{c|}{Same} & \multicolumn{4}{c|}{Unseen} & \multicolumn{4}{c}{Noise} \\
						\cmidrule{2-3}   \cmidrule{4-5} \cmidrule{6-7}   \cmidrule{8-9} \cmidrule{10-11}   \cmidrule{12-13} \cmidrule{14-15}
						&MAE(\textbf{\textit{R}})&MAE(\textbf{\textit{t}})&MIE(\textbf{\textit{R}})&MIE(\textbf{\textit{t}})&MAE(\textbf{\textit{R}})&MAE(\textbf{\textit{t}})&MIE(\textbf{\textit{R}})&MIE(\textbf{\textit{t}})
						&MAE(\textbf{\textit{R}})&MAE(\textbf{\textit{t}})&MIE(\textbf{\textit{R}})&MIE(\textbf{\textit{t}})\\
						\midrule
						ICP $\left(\star\right)$& 3.4339& 0.0114& 6.7706& 0.0227 &  3.6099&  0.0116&  7.0556& 0.0228& 4.6441&  0.0167& 9.2194& 0.0333  \\
						Go-ICP $\left(\star\right)$&3.5108 & 0.0115&6.8864  &0.0229& 3.5488&  0.0114&  6.9290& 0.0224& 3.1737&  0.0148&  5.7444& 0.0284  \\
						Super4PCS $\left(\star\right)$&1.5764 & 0.0035 & 2.7585 & 0.0069 &1.4562&  0.0034&  2.4281&  0.0067& 5.8263&  0.0160& 10.4796&  0.0317 \\
						FGR $\left(\star\right)$&  0.5972&  0.0021& 1.1563& 0.0041&  0.4579& 0.0016&  0.8442&  0.0032&  1.0676& 0.0036&  2.0038& 0.0072\\
						RANSAC $\left(\star\right)$&0.7031&0.0025 &1.2772 &0.0050&0.4427&0.0021 & 0.9447&0.0043&1.4316&0.0061 & 2.5345& 0.0120  \\
						IDAM $\left(\circ\right)$&  0.4243& 0.0020& 0.8170&  0.0040 & 0.4809& 0.0028& 0.9157& 0.0055& 2.3076& 0.0124& 4.5332& 0.0246\\
						RPM-Net $\left(\circ\right)$& 0.0051& \textbf{0.0000}& \textbf{0.0201} &\textbf{0.0000}&0.0064 &0.0001 & \textbf{0.0207}  &\textbf{0.0001}&0.0075 & \textbf{0.0000}& \textbf{0.0221} &\textbf{0.0001}\\
						FMR $\left(\triangleleft\right)$& 3.6497&  0.0101& 7.2810& 0.0200& 3.8594& 0.0114&7.6450& 0.0225& 18.0355& 0.0536& 35.7986& 0.1063 \\
						CEMNet $\left(\triangleleft\right)$ &0.1385 &0.0001 & 0.2489 & 0.0002 &0.0804 &0.0002 & 0.1405 & 0.0003 &10.7026 & 0.0393& 21.1836 & 0.0781\\
						\midrule
						Ours $\left(\triangleleft\right)$& \textbf{0.0033}& \textbf{0.0000}& 0.0210&\textbf{0.0000}& \textbf{0.0059}& \textbf{0.0000}& 0.0228& \textbf{0.0001} & \textbf{0.0069}&0.0001& 0.0230& \textbf{0.0001}\\
						\bottomrule
				\end{tabular}}
			}
		\end{center}
		\caption{The registration results of different methods on ModelNet40. $\left(\star\right)$, $\left(\triangleleft\right)$, and $\left(\circ\right)$ denote the traditional, unsupervised and supervised methods, respectively.}
		\label{table1}
		\vskip -5pt
	\end{table*}

	\begin{table*}
		\setlength{\abovecaptionskip}{0pt}
		\setlength{\belowcaptionskip}{0pt}
		\begin{center}
			{   \resizebox{1\linewidth}{!}{
					\begin{tabular}{lcccc|cccc|cccccccc}
						\toprule
						\multicolumn{1}{c}{\multirow{2}{*}{Model}} &\multicolumn{4}{c|}{ICL-NUIM} & \multicolumn{4}{c|}{7Scenes} & \multicolumn{4}{c}{KITTI} \\
						\cmidrule{2-3}   \cmidrule{4-5} \cmidrule{6-7}   \cmidrule{8-9} \cmidrule{10-11}   \cmidrule{12-13} \cmidrule{14-15}
						&MAE(\textbf{\textit{R}})&MAE(\textbf{\textit{t}})&MIE(\textbf{\textit{R}})&MIE(\textbf{\textit{t}})&MAE(\textbf{\textit{R}})&MAE(\textbf{\textit{t}})&MIE(\textbf{\textit{R}})&MIE(\textbf{\textit{t}})
						&MAE(\textbf{\textit{R}})&MAE(\textbf{\textit{t}})&MIE(\textbf{\textit{R}})&MIE(\textbf{\textit{t}})\\
						\midrule
						ICP $\left(\star\right)$ & 2.4022& 0.0699& 4.4832& 0.1410  & 6.0091& 0.0130& 13.0484&0.0260 & 4.7433& 0.9174& 11.9982&2.5742\\
						Go-ICP $\left(\star\right)$ &0.7339& 0.0259&1.3927&0.0522&7.0968& 0.0136&14.9701&0.0274& 2.3807&0.0431& 5.7684 & 0.0963\\ 
						Super4PCS $\left(\star\right)$ &1.3964&0.0547&2.7792&0.1103&1.6567&0.0081&2.9388&0.0162&2.2049&0.0384&5.2036&0.0849\\
						FGR $\left(\star\right)$ & 2.2477&0.0808& 4.1850&0.1573&0.0919&0.0004&0.1705&0.0008 &1.6777 & 0.0352& 4.0467&0.0762 \\
						RANSAC$\left(\star\right)$ &1.2349&0.0429 & 2.3167&0.0839&1.2325& 0.0062& 2.1875& 0.0124& 1.9353& 0.0230& 4.2766&0.0470\\
						IDAM$\left(\circ\right)$ &4.4153& 0.1385& 8.6178&0.2756&5.6727&0.0303&11.5949&0.0629 & 1.6348 & 0.0230 &3.8151 & 0.0491\\
						RPM-Net $\left(\circ\right)$ &0.3267 & 0.0125& 0.6277 &0.0246&0.3885 &0.0021 & 0.7649 &0.0042&0.9164 &\textbf{0.0146} &2.1291 &\textbf{0.0303}\\
						FMR$\left(\triangleleft\right)$ & 1.1085& 0.0398&2.1323&0.0786& 2.5438&0.0072&4.9089&0.0150& 1.6786&0.0329 &4.0571 & 0.0703\\
						CEMNet $\left(\triangleleft\right)$ &0.2374&\textbf{0.0005}& 0.3987 & \textbf{0.0010} &0.0559&0.0001&0.0772&0.0003&-&-&-&-\\ 
						\midrule 
						Ours $\left(\triangleleft\right)$ &\textbf{0.0492}& 0.0023&\textbf{0.0897}& 0.0049& \textbf{0.0121}& \textbf{0.0001}& \textbf{0.0299}& \textbf{0.0001}& \textbf{0.8251}& 0.0183& \textbf{1.8754}& 0.0414\\
						\bottomrule
				\end{tabular}}
			}
		\end{center}
		\caption{The registration results of different methods on Indoor Scenes and KITTI dataset. $\left(\star\right)$, $\left(\triangleleft\right)$, and $\left(\circ\right)$ denote the traditional, unsupervised and supervised methods, respectively. Besides, CEMNet does not provide the results on KITTI.}
		\label{table9}
		\vskip -5pt
	\end{table*}
	
	\section{Experiments}
	\subsection{Experimental Settings}
	\noindent \textbf{Datasets.} We evaluate our method on ModelNet40 \cite{wu20153d}, 7Scenes \cite{shotton2013scene}, ICL-NUIM \cite{choi2015robust} and KITTI odometry datasets \cite{geiger2012we}. The ModelNet40 consists of 12,311 meshed CAD models from 40 categories. We use 9,843 models for training and 2,468 models for testing. 7Scenes is a widely used benchmark registration dataset of indoor environment with 7 scenes including Chess, Fires, Heads, Office, Pumpkin, RedKitchen and Stairs.
	The dataset is divided into 296 samples for training and 57 for testing. 
	For another synthetic indoor scene ICL-NUIM, we first augment the dataset, then split the dataset into 1,278 samples for training and 200 samples for testing. And the KITTI odometry dataset consists of 11 sequences with ground truth pose, we use Sequence 00-05 for training, 06-07 for validation, and 08-10 for testing. We form the pairwise point clouds with the current frame and the 10th frame after it.
	
	\noindent \textbf{Compared methods and evaluation metrics.} We compare with traditional methods including ICP \cite{besl1992method}, Go-ICP \cite{yang2013go}, FGR \cite{zhou2016fast}, FPFH + RANSAC~\cite{fischler1981random}, and Super4PCS \cite{mellado2014super}. Besides, we compare with deep methods, including supervised IDAM \cite{li2020iterative} and RPM-Net \cite{yew2020rpm}, and unsupervised FMR \cite{huang2020feature} and CEMNet \cite{jiang2021CEMNet}. 
	We evaluate the registration by the mean absolute errors (MAE) of $\mathbf{R}$ and $\mathbf{t}$ used in DCP \cite{wang2019deep}, which are anisotropic, and the mean isotropic errors (MIE) of $\mathbf{R}$ and $\mathbf{t}$ used in RPM-Net.
	
	\noindent \textbf{Implementation details.} Our model is implemented in Pytorch. We optimize the parameters with the ADAM optimizer. The initial learning rate is 0.001. For ModelNet40 and KITTI, we train the network for 50 epochs and multiply the learning rate by 0.7 at epoch 25. For indoor scenes, we multiply the learning rate by 0.7 at epochs 25, 50, 75 and train the network for 100 epochs.
	
	\subsection{Comparison Evaluation on ModelNet40}
\begin{figure*}[t]
		\centering
		\includegraphics[width=\linewidth]{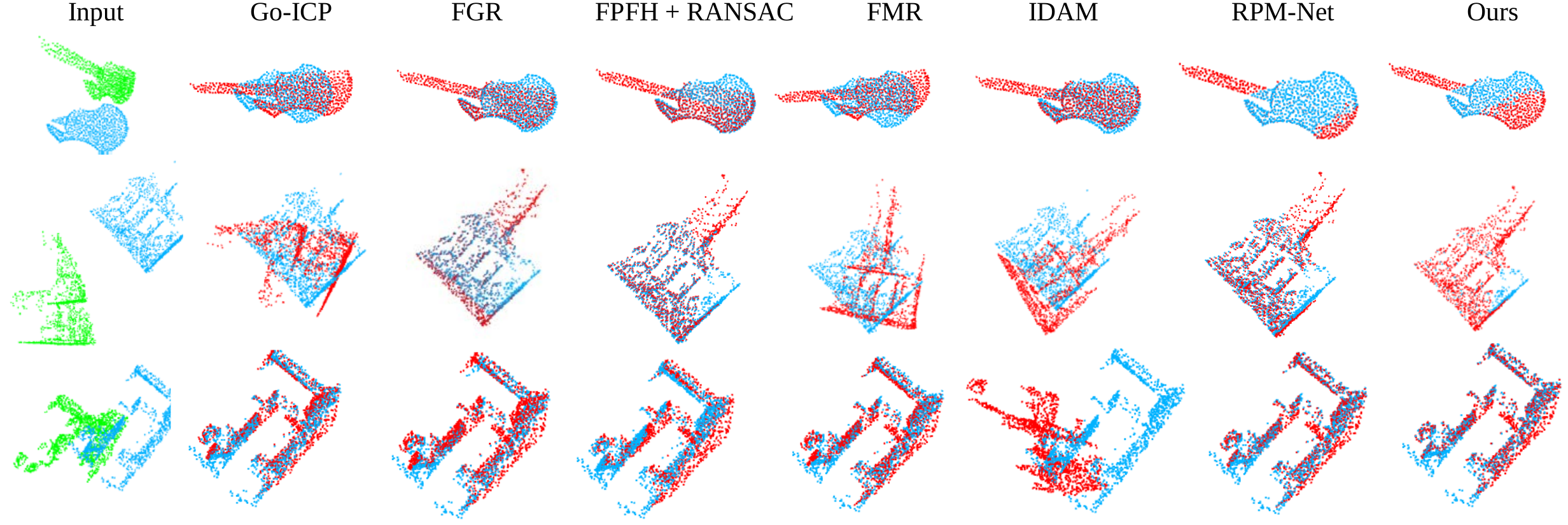}
		\caption{\textbf{Green}: source point clouds. \textbf{Blue}: target point clouds. \textbf{Red}: registration results. Visualization of registration results on the ModelNet40, 7Scenes and ICL-NUIM datasets.
		}
		\label{fig5}
		\vskip -5pt
	\end{figure*}

	\noindent \textbf{Same categories.}
	Each source point cloud $\mathbf{P}$ in the training set is transformed into $\mathbf{Q}$ by a randomly generated transformation matrix. The rotation along each axis is uniformly sampled in $[0,\ang{45}]$ and the translation along each axis is $[-0.5,0.5]$. To simulate partial-to-partial registration, we follow PRNet \cite{wang2019prnet} to remove $25\%$ points from both point clouds.
	From the left column of Table \ref{table1}, one can see that our method can obtain the lowest error among the traditional and unsupervised methods. Besides, our method can even outperform the supervised IDAM by a large margin. And we can achieve comparable results to the supervised RPM-Net.
	Some visualization results are presented in Figure~\ref{fig5}.
	
	\noindent \textbf{Unseen categories.} In order to demonstrate the effectiveness of our method, we train our model and all compared deep learning methods on 20 categories and test all methods on the other 20 categories. The test categories have not been seen during training. From the middle part of Table \ref{table1}, one can see that our method can obtain good performance in terms of the unseen point cloud registration.
	
	\noindent  \textbf{Gaussian noise.} In this experiment, we train the model on the noise-free data of ModelNet40 and test on a new test set with Gaussian noise. We generate Gaussian noise with $\mu = 0$ and $\sigma = 0.5$, and then clip the noise to $[-1.0,1.0]$. From the results listed in the right part of Table \ref{table1}, one can see that due to the neighborhood consensus based inlier selection, our unsupervised model is robust to noise. The global feature based unsupervised FMR is sensitive to noise.
	
	\subsection{Comparison Evaluation on Indoor/Outdoor Scenes}
	We further conduct comparison evaluation on two indoor scenes: synthetic ICL-NUIM and real-world 7Scenes. We resample the source point clouds to 2,048 points and operate the rigid transformation on them for the target point clouds, then downsample the point clouds to 1,536 points to generate the partial data. For outdoor KITTI dataset, we first augment the datasets with random rotations based on the initial pose. Then we voxelize the point clouds with 0.3m voxel size and randomly sample 2048 points, which are used for deep learning based methods. For traditional methods, we list the results of them with the original point clouds (We also test them with the preprocessed data, but the performance all deteriorates). From Table~\ref{table9}, one can see that our model obtains satisfactory performance on both indoor and outdoor scenes. The translation errors of supervised RPM-Net on KITTI are smaller than ours, but the runtime of our method is almost 1/2 of it. Besides, we achieve better results compared to FPFH + RANSAC with 1/8 runtime of it.
	
	\subsection{Ablation Study}\label{ablation}
	\noindent \textbf {The effectiveness of key components.}
	We conduct ablation study on two key components of our model: Matching Map Refinement (MMR) and Inlier Evaluation (IE) modules. In our baseline setting (BS), we concatenate the coordinates of predicted matching pairs as the inputs of MLPs to replace our IE module and remove the MMR module. The baseline model is trained with our unsupervised loss function. We add the components one by one to obtain their results of data in the cases of Gaussian noise and $50\%$ missing points on ModelNet40. The results can be seen in Table~\ref{table4}. One can see that inlier evaluation module can outperform the baseline by a large margin. In addition, the matching map refinement module based on neighborhood consensus can contribute to obtain more accurate registration results of noisy data.
	
	\begin{table}[t]
		\setlength{\abovecaptionskip}{0pt}
		\setlength{\belowcaptionskip}{0pt}
		\begin{center}
			\resizebox{1\linewidth}{!}{
				\begin{tabular}{lcccc}
					\toprule
					\multicolumn{1}{c}{Model}&MAE(\textbf{\textit{R}})&MAE(\textbf{\textit{t}})&MIE(\textbf{\textit{R}})&MIE(\textbf{\textit{t}})\\
					\midrule
					BS &2.2207 &0.0319&4.3694&0.0632\\ 
					BS + IE& 0.1638&0.0007&0.2832&0.0014\\ 
					BS + IE + MMR   &\textbf{0.0266}& \textbf{0.0003}& \textbf{0.0504}& \textbf{0.0005}\\
					\bottomrule
			\end{tabular}}
		\end{center}
		\caption{The results of different combination of key components on ModelNet40.}
		\label{table4}
		\vskip -5pt
	\end{table}
	
	\begin{table}[t]
		\setlength{\abovecaptionskip}{0pt}
		\setlength{\belowcaptionskip}{0pt}
		\begin{center}
			\resizebox{1\linewidth}{!}{
				\begin{tabular}{lcccc}
					\toprule
					\multicolumn{1}{c}{Loss}&MAE(\textbf{\textit{R}})&MAE(\textbf{\textit{t}})&MIE(\textbf{\textit{R}})&MIE(\textbf{\textit{t}})\\
					\midrule
					GAL &0.6303&0.0095&1.1601&0.0191\\ 
					GAL + SCL &0.1165&0.0020&0.2023&0.0039\\ 
					GAL + SCL + NCL  &\textbf{0.0121}&\textbf{0.0001}&\textbf{0.0299}&\textbf{0.0001}\\
					\bottomrule
			\end{tabular}}
		\end{center}
		\caption{The results of different combination of loss functions on 7Scenes.}
		\label{loss}
		\vskip -5pt
	\end{table}
	
	\noindent \textbf{Loss functions.}
	In our experiments, we train our model with the combination of the Global Alignment loss (GAL), the Neighborhood Consensus loss (NCL), and the Spatial Consistency loss (SCL). In Table \ref{loss}, we train our model using the different loss functions and present the results on 7Scenes. One can see that with the neighborhood consensus loss and spatial consistency loss based on the selected inliers, we can obtain high registration precision.
	
	\noindent \textbf {Matching map refinement module.} Directly adding the neighborhood-wise matching map $\mathbf{S}$ with point-wise matching map $\mathbf{M}$ decreases the discriminability of points in the same neighborhoods. The MAEs are 0.0072, 0.0002, and MIEs are 0.0338, 0.0002. Compared with the left column of Table \ref{table1}, one can see that our proposed matching map refinement module can yield better results.
	
	\noindent \textbf{Inlier evaluation module.}
	From Table \ref{table4}, one can see that the results of directly concatenating the putative correspondences for inlier evaluation are worse than our designed module, which ignores matching validity from neighborhood consensus. Besides, we also train our model without enhancing the message propagation between adjacent points via a large convolutional kernel ($1 \times 3$), we replace it with $1 \times 1$ kernel. The MAEs are 0.0294 and 0.0001, the MIEs are 0.0590 and 0.0002 on data with Gaussian noise. Compared with the results on the right column of Table \ref{table1}, one can see that enhancing the message propagation in the graphs can obtain better results with noisy data.
	
	\noindent \textbf {Feature update on the transformed source point clouds.} Since the features extracted by DGCNN are not fully rotation invariant, updating the features of transformed source point clouds which are gradually aligned to target point clouds along with the iterations will generate more accurate correspondences. Without updating the features, for samples from [\ang{45}, \ang{90}] on ModelNet40, the MAEs are 1.4653 and 0.0046, the MIEs are 2.0210 and 0.0094. As contrast, the MAEs of updating the features are 0.4213 and 0.0014, the MIEs are 0.7116 and 0.0027. One can see that it is necessary to update the features of transformed source point clouds along with iterations.
	
	\begin{figure}
		\centering
		\includegraphics[width=\linewidth]{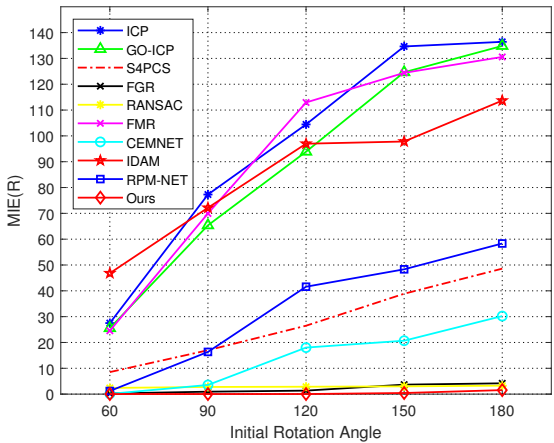}
		\vskip -10pt
		\caption{The rotation errors of different initial rotation angles on the 7Scenes.}
		\label{rotation}
		\vskip -5pt
	\end{figure}
	
	\noindent \textbf {Robustness to large rotations.}
	In order to demonstrate the robustness to large rotations, we utilize the samples from [\ang{0}, \ang{45}] as training and samples from [\ang{60}, \ang{180}] as testing. The rotation errors of different methods are shown in Figure \ref{rotation}. One can see that our model only with the xyz coordinates as the inputs is robust to large transformations, we achieve \ang{1.6} rotation error of \ang{180} initial rotation. Besides, FGR and FPFH + RANSAC rely on the rotation invariance FPFH, so the performance does not degrade dramatically.
	
	
	
	\noindent \textbf{Inference time.} We calculate the inference time with an Intel I5-8400 CPU and Geforce RTX 2080Ti GPU. The running time is measured in millisecond with a batch size of 1, averaged over the entire test set of ModelNet40. We provide the timings for 3 iterations as used in previous experiments for our method. Note that the classical methods are executed on CPU, but the learning based methods on a GPU. The running time are 8 (ICP), 41 (IDAM), 52 (FGR), 57 (Ours), 82 (FPFH+RANSAC), 86 (RPM-Net), 310 (CEMNet), 386 (FMR), 2740 (Go-ICP), and 6129 (Super4PCS). 
	
	\section{Conclusion}
	We proposed an end-to-end unsupervised point cloud registration framework based on neighborhood consensus for effective inlier evaluation. Specifically, we developed a neighborhood consensus based matching map refinement module to obtain a more accurate matching map and generated high-quality matching points to form a pseudo target point cloud. With the source point cloud and corresponding pseudo target point cloud, we designed the inlier evaluation module to distinguish inliers via capturing the geometric difference between the source neighborhoods and the corresponding pseudo target neighborhoods.
	Finally, we formulated our combined loss to train our network. Extensive experiments on the ModelNet40, ICL-NUIM, 7Scenes and KITTI benchmarks demonstrate that our unsupervised framework can achieve outstanding performance.
	\section{Acknowledgements}
	This work was supported by the National Science Foundation of China (Grant Nos. U1713208, 61876084).
	{
		\bibliographystyle{IEEEtran}
		\bibliography{aaai22}
	}
\end{document}